\documentclass[runningheads]{llncs}
\usepackage{graphicx}
\usepackage{amsmath, amssymb}
\usepackage{caption}
\usepackage{subcaption}
\usepackage{url}
\usepackage{xcolor}
\usepackage[normalem]{ulem}
\usepackage{comment}
\usepackage[mathscr]{euscript}
\usepackage[]{algorithm2e}
\usepackage{rotating}
\usepackage[T1]{fontenc}

\newcommand{\set}[1]{\mathscr{#1}}

\newcounter{applyCorrections}
\newcommand{\rimuovi}[1]{\ifthenelse{\value{applyCorrections}<1}{\textcolor{red}{\sout{#1}}}{}}
\newcommand{\aggiungi}[1]{\ifthenelse{\value{applyCorrections}<1}{\textcolor{blue}{#1}}{}}

\newcommand{\replace}[2]{\ifthenelse{\value{applyCorrections}<1}{\textcolor{red}{\sout{#1}} \textcolor{blue}{\emph{#2}}}{\textcolor{black}{#2}}}

\begin{document}

\title{Play Everywhere: A Temporal Logic based Game Environment Independent Approach\\for Playing Soccer with Robots}


%
\titlerunning{Play Soccer Everywhere with Robots using Temporal Logic}
%
\author{
V. Suriani \inst{1}  
 \orcidID{0000-0003-1199-8358} 
\and E. Musumeci \inst{1} \orcidID{0009-0004-2359-5032}
\and \\ D. Nardi \inst{1}\orcidID{0000-0001-6606-200X}
\and D. D. Bloisi \inst{2}\orcidID{0000-0003-0339-8651}
}
\authorrunning{Suriani et al.}
%
\institute{Dept. of Computer, Control, and Management Engineering\\ Sapienza University of Rome, Rome (Italy),
    \email{\{lastname\}@diag.uniroma1.it.} \and
Dept. of Mathematics, Computer Science, and Economics,\\University of Basilicata, Potenza (Italy),
\email{domenico.bloisi@unibas.it}
}
\maketitle              
\begingroup\renewcommand\thefootnote{\textsection}
\endgroup
\begin{abstract}


Robots playing soccer often rely on hard-coded behaviors that struggle to generalize when the game environment change. In this paper, we propose a temporal logic based approach that allows robots' behaviors and goals to adapt to the semantics of the environment. In particular, we present a hierarchical representation of soccer in which the robot selects the level of operation based on the perceived semantic characteristics of the environment, thus modifying dynamically the set of rules and goals to apply. 
The proposed approach enables the robot to operate in unstructured environments, just as it happens when humans go from soccer played on an official field to soccer played on a street.
Three different use cases set in different scenarios are presented to demonstrate the effectiveness of the proposed approach.



\keywords{semantic mapping \and multi-agent planning \and robot soccer}
\end{abstract}
%
%
%
\section{Introduction}

\begin{figure} [t]
\centering
\includegraphics[width=0.99\textwidth]{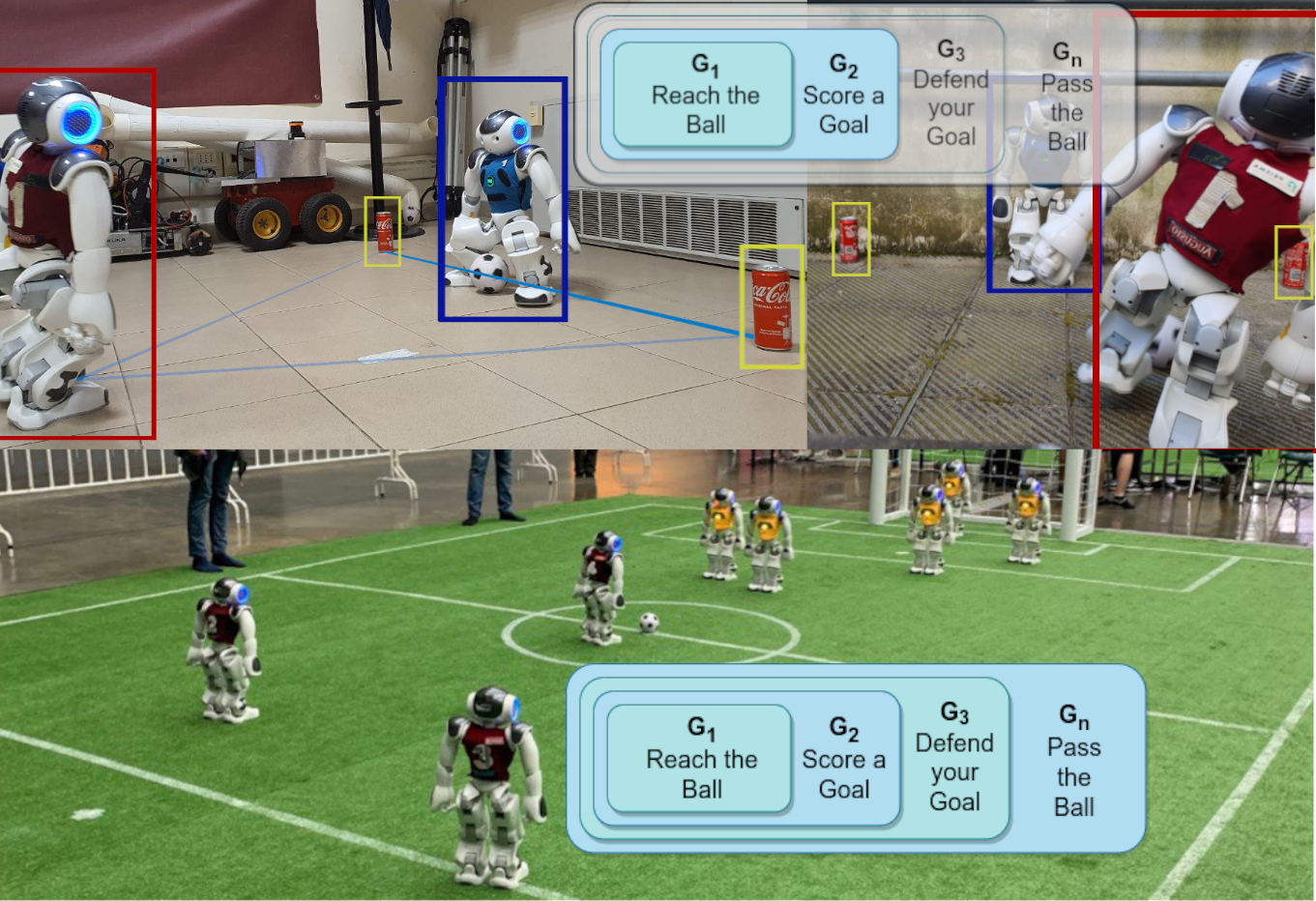}
\caption{Different scenarios where robots can play using the same architecture. The agent establishes the goal for the task from the semantics of the environment.} \label{fig:intro}
\end{figure}

The development of robots that can perform complex tasks in dynamic and unstructured environments is an open research problem. One of the challenges of this research area is to design control architectures that can adapt to changing conditions and generalize to new and unseen situations.

In this paper, we focus on autonomous robots playing soccer. In the domain of robot soccer, the majority of existing approaches rely on hard-coded behaviors that are specifically designed for certain game scenarios (e.g., using green carpets in indoor environments). While these methods can be effective in well-structured environments, they often fail to cope with unexpected situations or variations in the game conditions. Moreover, they are typically difficult to transfer to other domains or tasks, as they rely on a fixed set of rules and assumptions.

To address the above listed limitations, we propose a dynamic approach to robotic soccer, in which the robot derives the rules of the game from the semantics of the playing environment. Specifically, we suggest a hierarchical representation of soccer that allows the robot to choose the level of operation based on the perceived characteristics of the game scenario. In this way, the robot can select the appropriate set of rules to follow and dynamically modify its goal function accordingly.
Our approach enables the robot to operate in unstructured and dynamic environments, such as street soccer fields (see Fig. \ref{fig:intro}), where the traditional hard-coded approaches fail. Furthermore, our method is highly adaptable to different domains and tasks, as it does not rely on fixed rules or assumptions.

In the context of the RoboCup, adapting robot strategy to the environment can be hard to achieve given the fact that, in most leagues, state-machine behaviors are still predominant \cite{gamestrategies}, and, even when the deployed behavior is learning-based, the resulting policies can suffer in challenging and dynamic environmental conditions.

To enable a robot to operate in unstructured environments, we propose a hierarchical representation of the rules of the RoboCup agent controlled by a planning system capable of accepting constraints in real-time from the external environment. We believe that creating a mechanism for deploying robotic agents in environments where not all the semantic elements are constituted is preparatory to deal with the RoboCup 2050 challenge.

Our goal is to create a software architecture that can be used in every scenario, spanning from simple (i.e., a robot playing alone) to complex scenarios (i.e., two teams playing in a regular field using a specific set of rules).
In particular, to exploit the task adaptation capabilities in RoboCup SPL, we present an architecture to deduce the goal of the robotics system while preserving the full capability of the software, based on a planned strategy. 
In particular, we deployed online conditioning through Pure-Past Linear Temporal Logic (PPLTL) rules on finite traces, also known as \textit{PLTL$_f$} rules, expressing temporally extended goals and non-Markovian properties over traces, natively.
Our application allows for real-time generation of non-deterministic policies that are able to fit different situations in which robots have to spot the semantic relevant elements in an environment and play using a goal deduced from them, as shown in Fig. \ref{fig:intro}.


The contribution of this work is three-fold.
\begin{enumerate}
    \item A hierarchical representation of the goal capable to adapt to the complexity level of the operational environment, based on a measure of the perceived semantics that can be extracted from the robot sensors.
    \item An SPL-related conceptual layer capable to generalize and to map ordinary objects to semantically equivalent entities, decoupled from the sensory level.
    \item An approach to FOND planning for temporal goals, enabling the system to model the uncertainty of the environment and to manage such uncertainties, by encoding non-Markovian properties on the resulting finite traces, as mixed sets of predicates and $PLTL_f$ rules.
\end{enumerate}   

The rest of the paper is organized as follows. In Section \ref{sec:relwork}, we survey the current state-of-the-art; in Section \ref{sec:back}, we expose a brief theoretical background to introduce the concepts expressed in the paper; in Section \ref{sec:method}, we show more in detail the proposed method; in Section \ref{sec:examples}, we illustrate three use cases of the presented system. Finally, in Section \ref{sec:conclusion}, we draw the conclusions and the possible future developments of this work.


\section{Related Work}
\label{sec:relwork}



Traditional approaches to programming robot soccer behavior often rely on hard-coded rules that struggle to generalize when the game location changes \cite{gamestrategies}. To address this issue, recent research has explored the use of machine learning techniques to develop more adaptive robot soccer players by creating new environments based on real soccer scenarios as in \cite{kurach2020google}.

However, these approaches still are leading to very specific robotic behaviors and policies capable to behave properly in a structured scenario with a standardized set of elements and a fixed set of rules.
We propose a dynamic approach in which the rules of the game are derived from the semantics of the environment and extracted from the full standard set of rules. 

To let the robots deal with the semantics of the environments, in \cite{pronobis2012large} a semantic mapping system is divided into four layers on three levels, spanning from a sensory layer, to a categorical and a place layers is presented. In \cite{bastianelli-bloisi-gemignani}, an online approach enables online semantic mapping, that permits adding elements acquired in long-term interaction with the users to the representation. None of these approaches tacked a peculiar field as the robotic soccer.

Our approach is specifically tailored to the game of soccer, and we present a hierarchical representation of soccer in which the robot selects the level of operation based on the perceived characteristics of the environment. The hierarchical representation helps to scale in different scenarios, adapting to new goals. \replace{In fact, when the environment becomes complex}{In place of temporal constraints}, the online generation of plans can be time-consuming. To this end, \cite{de2018automata} proposes FOND Planning with Linear Temporal Logic over finite traces (LTL$_f$) \rimuovi{with temporally extended goals} in Fully Observable Non-Deterministic (FOND) domains. 
\cite{de2021fondpltlf} extends the previous approach to PLTL$_f$ goals by suitably modifying standard FOND planning domains.
In \cite{musumeci}, PLTLf \replace{has been adopted in order}{was used} to condition a RoboCup team's behavior based on high-level commands from a human coach, encoded as additional temporal constraints in the original goal for the FOND planning problem at hand (in this case, the task specific to the robot role in the multi-agent team).

Our approach differs from previous work because the set of PLTL$_f$ rules is \replace{dynamically}{automatically} chosen based on the perceived characteristics of the environment\rimuovi{ and not communicated by an external coach}.
\replace{Finally, we dynamically modify the goal function, which is dependent on the set of PLTL$_f$ rules,}{The PLTL$_f$ goal function is dynamically modified} to adapt to changing conditions \replace{. This approach enables}{enabling} the robot to operate in unstructured environments, \replace{such as a robot soccer player playing on}{like} a street soccer field, where the characteristics of the environment may change rapidly and require complex properties to be expressed without restructuring the whole planning domain. 
In summary, the proposed approach has the potential to enable more adaptive and robust robot players, and could also be applied to other domains where environmental cues can be used to determine robot behavior.

\section{Background}
\label{sec:back}
Among the tools we used to automatically scale the robot policy based on the environment semantics, we need to introduce planning with a PLTL$_f$ goal formula.
In \cite{de2013LTLsynthesis} and \cite{de2018automata}, a plan that entails an LTL goal formula is obtained by checking the non-emptiness of the product of a deterministic automaton (modeling the planning domain) and a non-deterministic automaton modeling the goal formula. In FOND planning, a DFA game is to be solved on said cross-product, to obtain a policy.

Following \cite{de2021fondpltlf}, given a pre-existing, non-deterministic planning domain $\set{D}$ and a PPLTL planning goal $G$, a new domain $D'$ and a new goal $G'$ can be compiled, so that any off-the-shelf non-deterministic planner can compute a plan for the new goal, treating the problem as a typical FOND planning problem. Given the set $\mathbf{sub}(G)$ of logical sub-formulae recursively obtained by the original goal $G$, a minimal subset $\Sigma_G \subseteq \mathbf{sub}(G)$ of sub-formulae of $G$ can be computed.\\
Then, given the a formula $g \in \Sigma_G$, a propositional interpretation $s_i$ over the set of predicates $\set{P}_{deduced}$ in the current planning domain, in the current state at instant $i$, and a specific interpretation $\sigma_i : \Sigma_G \rightarrow \{\top, \bot\}$, keeping track of the truth values of sub-formulas in $\Sigma_G$ in the current state at instant $i$, a predicate $\mathbf{val}(g, \sigma_i, s_i)$ can be defined for each sub-formula $g \in \mathbf{sub}(G)$ to track recursively the truth value of the PLTL$_f$ sub-formula $g$ at the current state.
 
 Given a trace $\tau$ over $\set{P}_{deduced}$, the satisfaction of the original goal $G$ is equivalent to the truth value in the final state of the predicate $\mathbf{val}(\cdot)$:
 \begin{equation}
 \label{eq:new_goal}
    \tau \models G \text{ iff } \mathbf{val}(G, \sigma_f, s_f)
 \end{equation}
 where $\sigma_f$ is the interpretation of $\Sigma_G$ in the final state and $s_f$ is the set of predicates of $\set{P}_{deduced}$ verified in the final state of the trace. At this point, each predicate $\mathbf{val}(g, \sigma_i, s_i)$ can be included as an additional propositional variable in the state of the planning domain, obtaining a new planning domain $\set{D'}$. According to equivalence \ref{eq:new_goal}, the new planning problem $\Gamma'$ will feature a new goal $G' = \mathbf{val}(G, \sigma_f, s_f)$.
 Given the original planning problem $\Gamma$ with a PLTL$_f$ goal $G$ and the new planning problem $\Gamma'$, featuring the new planning domain $\set{D'}$, with the new goal $G'$, then $\Gamma$ has a winning strategy $\pi$ if and only if $\Gamma'$ has a winning strategy $\pi'$.
 
\section{Proposed Approach}
\label{sec:method}

Our approach consists of three main components: 1) A perception module that extracts relevant features from the environment, 2) A hierarchical representation of the task (the soccer game), and 3) A decision-making module that selects the appropriate level of operation and goal function. To model the hierarchical representation and the decision-making module, we start from the definition of a \textit{Semantic Map} in the formalization introduced in \cite{Capobianco2015}, with some modifications. Specifically, we consider a representation composed by a tuple of three elements
\begin{equation}
    \set{SM} = \langle R, \set{M}, \set{P} \rangle
    \label{eq:semantic_map}
\end{equation}
where:
\begin{itemize}
    \item $R$ is the global reference system in which all the elements of the semantic map are expressed;
    \item $\set{M}$ is a set of geometrical elements obtained from sensor data, expressed in the reference frame $R$, describing spatial information in mathematical form. 
    \item $\set{P}$ is a set of predicates, describing the environment.
\end{itemize}

Here, the definition of a unique reference frame $R$ allows associating the elements of $\set{M}$ with those in
$\set{P}$. Then, given two semantic maps $\set{SM}_1 = \langle R_{1},\set{M}_1,\set{P}_1 \rangle$ and $\set{SM}_{m} = \langle R_{m},\set{M}_{m},\set{P}_{m} \rangle$ generated as an expected model of the environment, an evaluation metric can be defined as
\begin{equation}
  \delta(\set{SM}_1,\set{SM}_{m}) = f(|\set{M}_1 \ominus \set{M}_{m}|, |\set{P}_1 \boxminus \set{P}_{m}|),
  \label{eq:metrics}
\end{equation}
where $R_{1}$ and $R_{m}$ must coincide (e.g., through a simple geometrical transformation). 
It is important to notice that, since Eq. \ref{eq:metrics} assumes a unique reference frame, it is particularly suitable for the analysis of different semantic maps which are used to distinguish different semantic contexts in which the robot has to operate.
In fact, it allows reformulating the problem of semantic maps comparison as the problem of anchoring their representation to a common reference frame. In the deployed environment, objects and their semantic characterization have to belong to the same reference frame as the robotic agent maps them. 
Such metrics can be applied at multiple levels of abstraction, spanning from the low sensory level ($\ominus$) to a conceptual level ($\boxminus$), as presented in~\cite{Capobianco2015}.

The set $\set{M}$ is related to the environmental elements that can be perceived by the agent. 
In the case of the soccer environment, the set of field elements is finite and represented by the field elements:
\begin{equation}
    \set{M}_{m} = \{Ball, Goal Posts, Players, Field\}
\end{equation}
But it can also include referees, gestures, and any other element of the game that can be perceived by the agent.
The set of predicates contains the full set of rules of the game that can be relaxed. In fact, $\set{P}$ is derived by the set of rules of the soccer game using the semantic evaluation $\delta$ as a metric. Only the predicates referring to $\set{M}$ and coherent with the global $\set{R}$ are taken into account:
\begin{equation}
    \set{P(\delta)} \subseteq \set{P_{SPL}} 
\end{equation}

Being $P_i(\delta)$ the $i-th$ level in the rule selection hierarchy and indicating it as $\set{P}_i$ for clarity of notation, the subsequent goals hierarchy is expressed as 

\begin{equation}
    \set{G}_{m} = \bigcup_{i=1}^n G(\set{P}_i) \supseteq \bigcup_{i=1}^{n-1} G(\set{P}_i) \supseteq \cdots \supseteq G(\set{P}_1),
\end{equation} 
where $G_i$ represents the set of goals for the $i$-th level of the rules, and $n$ is the total number of stages.

As we progress through the inclusion of new rules, each set of goals (represented by $G_i$) builds upon the previous set of goals. The notation $\bigcup_{i=1}^n G_i$ represents the union of all the sets of goals up to the n-th stage, while $\bigcup_{i=1}^{n-1} G_i$ represents the union of all the sets of goals up to the (n-1)-th stage. 

Each set of goals is a subset of the next set of goals, with the final set of goals ($G_1$) being the most basic and foundational. For example, for a single robotic soccer player, if the only perceived element of the environment is the ball, the fundamental goal is to reach and hold the ball. Perceiving a set of goal posts, the agent's goal is extended in reaching and holding the ball and scoring to the goal. If in the semantics of the environment, the robot spots a couple of goals, the agent goal will be extended in order to take into account the presence of two teams and, hence, a goal that have to be defended and one that has to be used for scoring. Finally, in the case of perceived full robotic fields the robot refers to the full set of rules of the SPL.  

In the presented architecture, when dealing with the robot behavior, the $\set{SM}$ influences the policy generated for the FOND planning problem through the constraints expressed on the set of deduced predicates made available by the conceptual layer.
In fact, $\set{SM}$ conditions the behavior of the robot assembling a goal based on the set of rules that are related to items of the Semantic Map. 
\begin{equation}
   \pi = (U | S, \set{SM}) = (U | S, (\set{R}, \set{M}_RS, \set{P}_{deduced}))   
\end{equation}
$\set{P}_{deduced}$ is a subset of the planning domain $\set{D}$ predicates.
The FOND policy conditioned by the semantic mapping leads automatically to a new policy:
\begin{equation}
   \pi_{\bigcup_{i=1}^n G_i} = (U | S, \set{P}_{deduced})  
\end{equation}

This approach aims at the deployment of the robotic agent in an unstructured environment where the goal is defined by a $PLTL_f$ formula $\phi$ activated by the set of elements that the robot can perceive.

As shown in \cite{de2021fondpltlf},
$PLTLf$ goals are computationally advantageous with respect to $LTLf$ when temporal goal specifications are naturally expressed with respect to finite past traces. 
With those premises, the policy $\pi$ is the solution for a corresponding \textit{FOND} planning problem $\Gamma$ with a $PLTL_f$ goal defined in \cite{de2018automata} as
\begin{equation}
   \Gamma := \langle \set{D}, s_0, \phi \rangle  
\end{equation}
where $\set{D}$ is a FOND domain model, $s_0$ is the initial state and $\phi$ is a $PLTL_f$ formula. In the proposed approach, the equation above becomes
\begin{equation}
   \Gamma := \langle \set{D'}, s_0, \phi(\bigcup_{i=1}^n G_i) \rangle  
\end{equation}
where $\set{D'}$ is a planning domain where each sub-formula is represented by fluents, $s_0$ is the initial state common to all the traces of the robot behavior and the goal depends on the sum of the $n$ sets of goals $G_i$ deduced in the current context.

\subsection{Functional Architecture}
\label{sec:architecture}

\begin{figure} [t]
\centering
\includegraphics[width=1\textwidth]{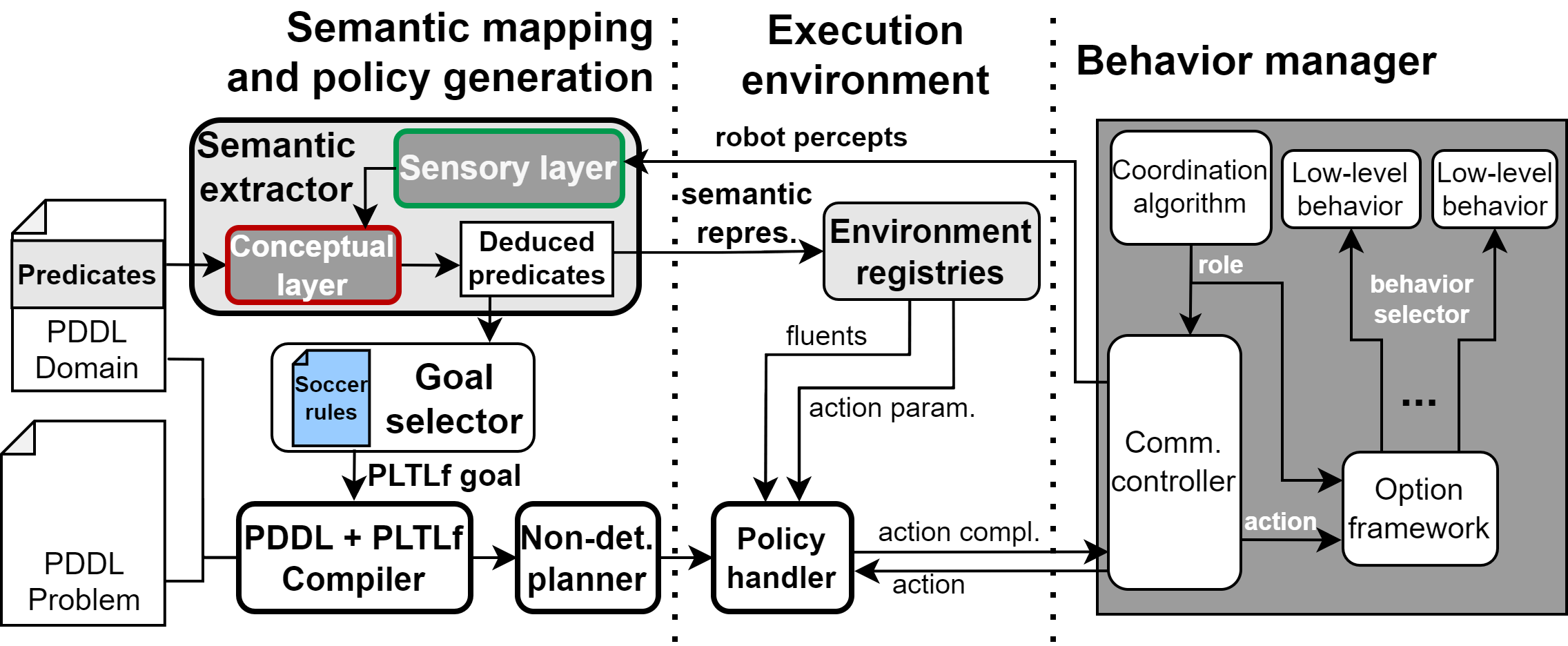}
\caption{Functional architecture. The \emph{sensory layer}, here highlighted in green, is shown in Fig. \ref{fig:sm_sensory_layer}, while the conceptual layer, in red, is shown in Fig. \ref{fig:sm_conceptual_layer}.} \label{fig:pipeline}
\end{figure}

Fig. \ref{fig:pipeline} shows the three main components of our architecture: 1) The perception module, which extracts relevant features from the environment, 2) the hierarchical and conceptual representation of the task (the soccer game), and 3) the decision-making module that selects the appropriate level of operation and goal function. The architecture relies on the B-Human Team SPL framework \cite{BHumanCodeRelease2021}. 

\begin{figure} [t]
\centering
\includegraphics[width=1\textwidth]{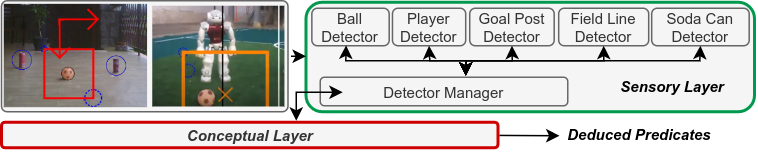}
\caption{The detection modules in the \emph{sensory layer} communicating with the conceptual layer, in red and shown in Fig. \ref{fig:sm_conceptual_layer}.} \label{fig:sm_sensory_layer}
\end{figure}

\paragraph{\textbf{Sensory Layer and Perception Modules.}}
The classification of the environment starts from the perception modules. We rely on different classifiers to percept the field elements of the environment in order to reactive the unused ones (see Fig. \ref{fig:sm_sensory_layer}). To be able to play in every scenario, we created classifiers able to detect common objects, easy to retrieve. In particular, for the use-cases presented, we used:
\begin{itemize}
    \item a \textit{Ball Perceptor} capable to detect the SPL ball on any background and under any light conditions, based on a OpenCV Haar Cascade Classifier.
    \item a \textit{Field Perceptor} to identify the official SPL field and its elements.
    \item a \textit{Goal Perceptor} capable to detect official SPL goal posts on any terrain.
    \item a \textit{Soda Can Perceptor} to perceive soda can as elements representing goal posts, developed with an OpenCV Haar Feature-based Cascade classifiers.
    \item a \textit{Player Perceptor}, Deep Learning based.
\end{itemize}
The perception module is responsible for extracting relevant features from the environment and providing them to the decision-making module.

\begin{figure} [t]
\centering
\includegraphics[width=1.0\textwidth]{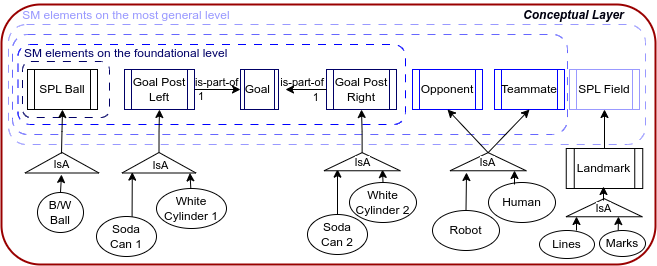}
\caption{Hierarchical architecture showing the relationships between the perceived objects and the semantic map in the conceptual layer.} \label{fig:sm_conceptual_layer}
\end{figure}

\paragraph{\textbf{Hierarchical representation and Conceptual Layer.}}
Fig. \ref{fig:sm_conceptual_layer} shows the hierarchical representation of the soccer game. It consists of multiple levels, each representing a possible abstraction of the game and a goal that depends on the semantic reconstruction of the environment. This is based on the conceptual layer, which compromises common-sense knowledge about concepts, and relations (mainly \textit{IsA} and \textit{IsPartOf} relations) between those concepts, and instances of spatial entities. The conceptual layer is built based on a pre-defined list of possible objects but it can be easily extended in order to include equivalences between entities belonging to official SPL fields and other heterogeneous environments. Here, the elements are hierarchically classified, and from this representation, the set of predicates extracted $\set{P}_{deduced}$ are used to condition the planner in the next module.

\paragraph{\textbf{Decision-making module using PLTLf goals over PDDL domains.}}
The uncertain set of features made available by the conceptual layer determines the level of operation of the agent and the planning goal that is dynamically adapted to it. The problem is modeled as a FOND planning problem in which the goal is formulated as a PLTL$_f$ formula using predicates in $P_{deduced}$ made available by the \emph{conceptual layer} of the Semantic Extractor. Following \cite{de2021fondpltlf}, this formalism allows to have generated traces satisfy temporal constraints, to model RoboCup SPL rules or the goal $G$ required at the current level of operation.

Given the goal $G$ and a pre-existing PDDL domain $\set{D}$, describing the actions that the agent can execute in the RoboCup SPL setting, with a set of predicates containing the set of deduced predicates $P \supseteq P_{deduced}$, we can obtain the new problem $\Gamma'$, featuring the new domain $\set{D'}$, such that any off-the-shelf planner will produce a policy that realizes the original goal $G$, by realizing a new goal $G'$ that is compatible with the original one.
 Following \cite{de2021fondpltlf}, the evaluation of the new goal $G'$ is equivalent to the evaluation of the predicate $\mathbf{val}(G, \sigma_f, s_f)$.

 For each predicate $g_i \in \mathbf{sub}(G)$, a new predicate $\mathbf{val}_{g_i} = \mathbf{val}(g_i, \sigma_i, s_i)$ is added to the planning domain $\set{D}$. The evaluation of all such propositions will depend on the evaluation of a reduced set of propositional predicates $g_i \subseteq \Sigma_G \subseteq \mathbf{sub}(G)$. Using some features currently well-supported by most off-the-shelf planners, the new set of propositional rules is embedded in the pre-existing PDDL planning domain. For each sub-formula $g \in \mathbf{sub}(G)$, a corresponding predicate $\mathbf{val}_g$ is embedded into the original PDDL domain as a \emph{derived predicate}, so that the result of its evaluation is equivalent to the result of $\mathbf{val}(G, \sigma_i, s_i)$.
 
 Domain actions are also modified accordingly: using the PDDL construct for \emph{conditional effects}, the effects of each domain action are populated with rules for the update of the new set of derived predicates, of the form:
 \begin{equation}
      \text{ }\mathbf{val}_g \text{ } \rightarrow \text{ } f(g) \in \Sigma_g
      \hspace{2.5cm}
     \mathbf{\neg val}_g \rightarrow \neg f(g) \in \Sigma_g
 \end{equation}
 where $f(g)$ is a propositional formula in the reduced set $\Sigma_g$ that is relevant for the evaluation of $\mathbf{val}_g$. These rules are added to the effect of all domain actions, so that the truth values of each $\mathbf{val}_g$ are updated every time an action is performed. As shown in \cite{de2021fondpltlf}, the size of the new planning problem is polynomial in the size of the original problem and, in particular, the number of additional predicates introduced is linear in the size of the PPLTL goal $G$.
 
 The resulting PDDL problem will have a new goal $G' = \{\mathbf{val}_G\}$. Given the original planning problem $\Gamma$ with a PLTL$_f$ goal $G$ and the new planning problem $\Gamma'$ with the new goal $G'$, then $\Gamma$ has a winning strategy $\pi$ if and only if $\Gamma'$ has a winning strategy $\pi'$.
 In this case, any sound and complete planner returns a policy $\pi'$ generating robot behaviors that are compliant with the original temporal goal.
 
In scenarios where fluents from the environment are needed to model conditions that are not known at planning time, "$oneof$" constructs are used in the post-conditions of actions to enumerate their possible unpredictable results, featuring predicates whose value will depend on the agent’s own world model and percepts at runtime. Policies generated with non-deterministic planners from the provided PDDL domain are therefore robust for a set of unpredictable outcomes.\\ The generated policy is not yet ready to be executed with temporally-extended actions. To overcome this problem, a pipeline composed by a set of "\emph{environment registries}", used to ground policy fluents, actions and their arguments, to elements made available by the conceptual layer, similarly to the architecture presented in \cite{musumeci}.

\section{Use-cases}

In order to test the presented approach, we propose three scenarios, in which the agent has to gather data from the environment, extract the set of elements belonging to the environment, determine the level of operation given the available elements, assemble the corresponding goal and compute a policy accordingly.

The three use-cases present a progressively increasing number of features made available by the conceptual layer, leading to an increasing level of complexity in the resulting goal, in the number of temporal constraints. The MyND planner \cite{mynd} was used for the presented experiments.
\label{sec:examples}

\begin{figure}[t]
     \centering
     \begin{subfigure}[t]{0.085\textwidth}
         \centering
         \includegraphics[width=\textwidth]{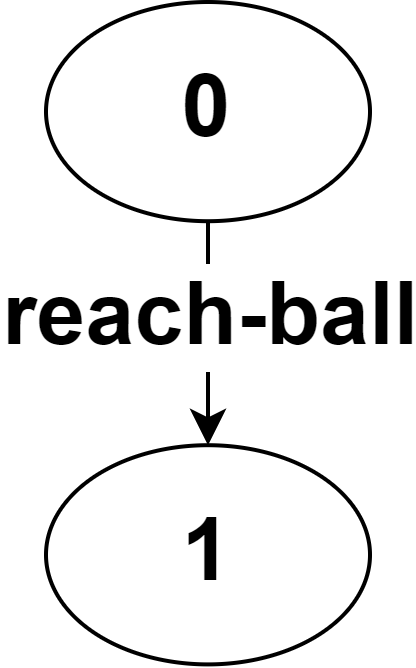}
         \caption{}
         \label{fig:policy1}
     \end{subfigure}
     \hfill
     \begin{subfigure}[t]{0.4\textwidth}
         \centering
         \includegraphics[width=\textwidth]{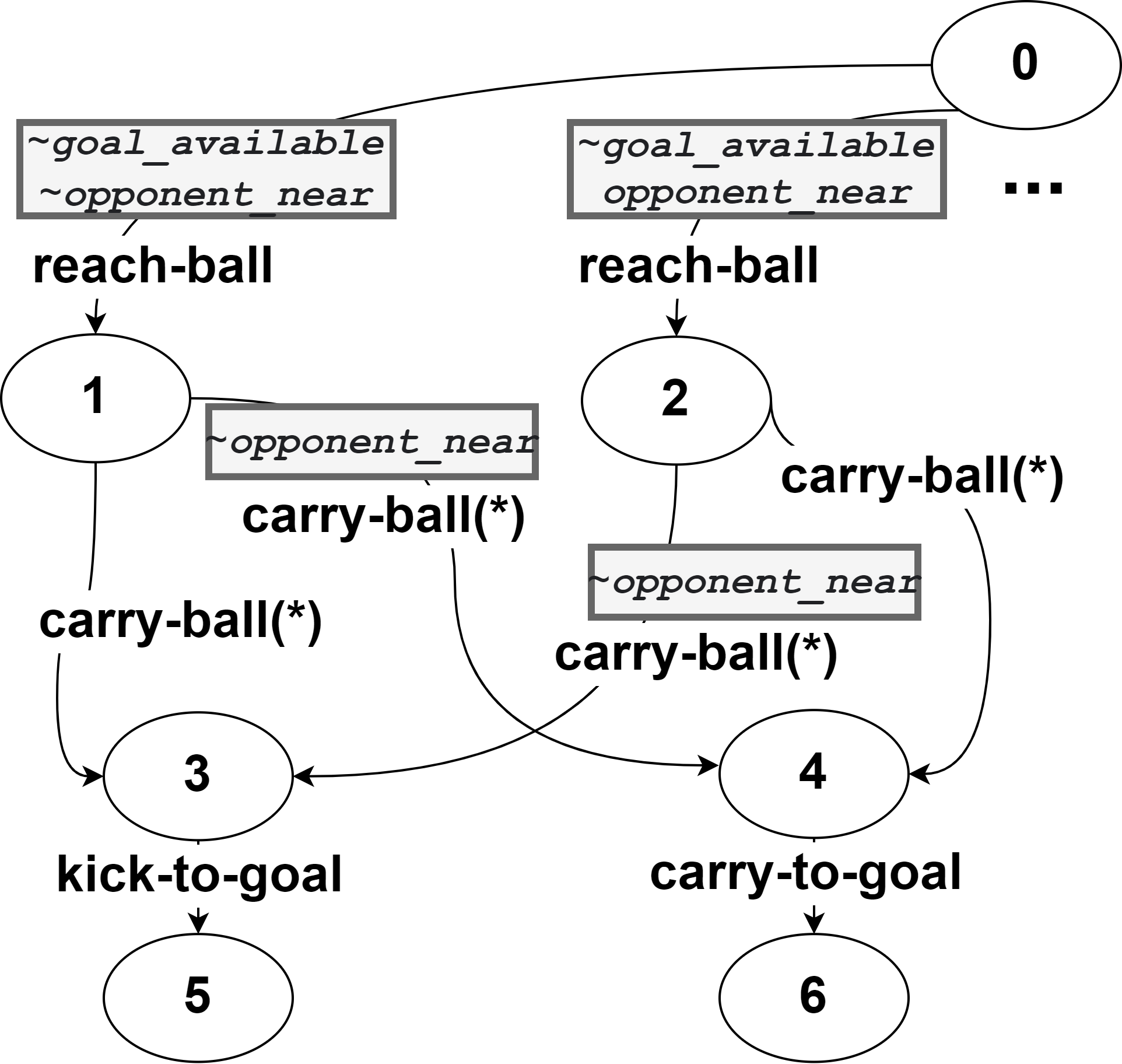}
         \caption{}
         \label{fig:policy2}
     \end{subfigure}
     \hfill
     \begin{subfigure}[t]{0.4\textwidth}
         \centering
         \includegraphics[width=\textwidth]{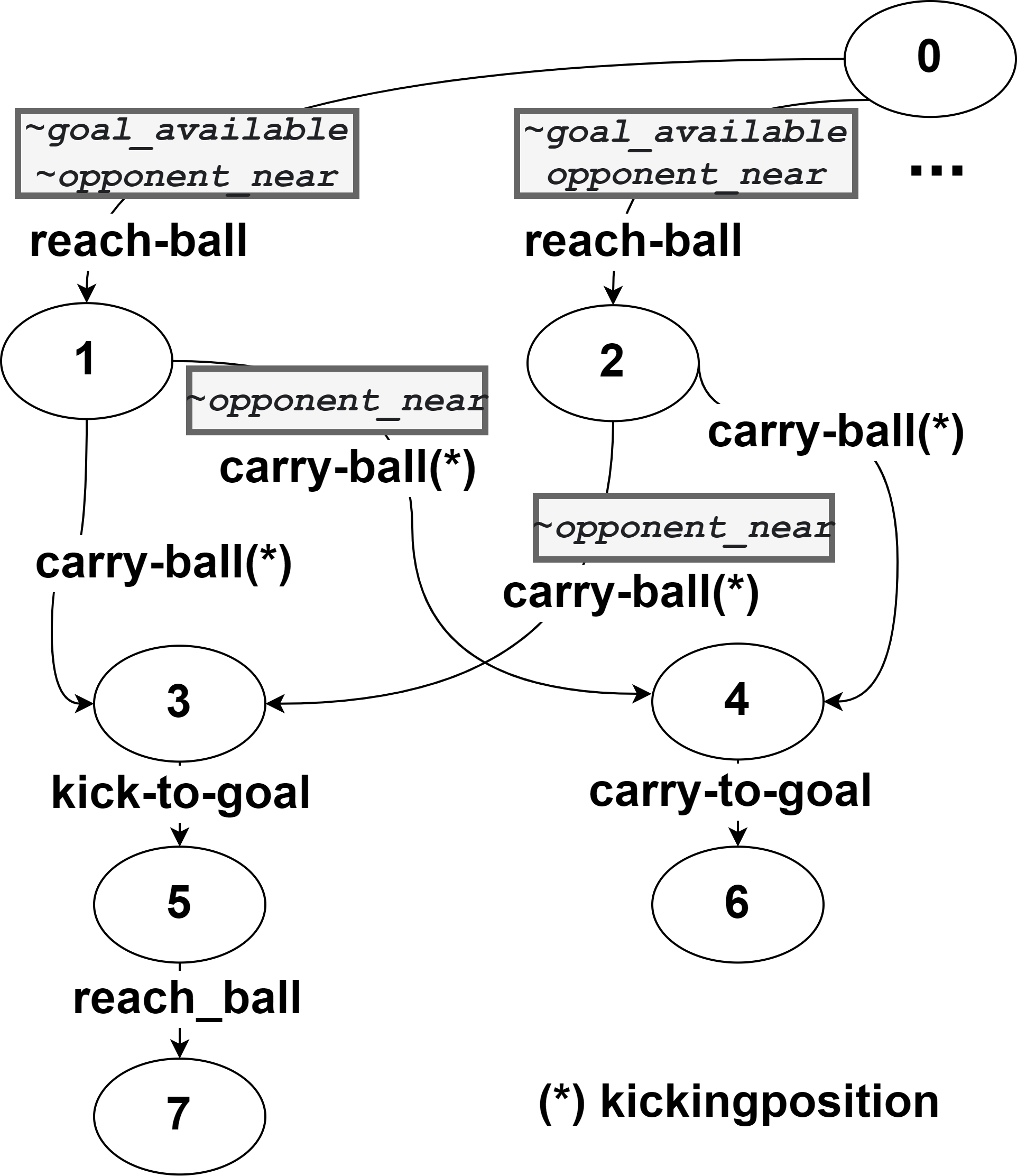}
         \caption{}
         \label{fig:policy3}
     \end{subfigure}
\caption{Policies generated from goals of increasing complexity. 
From left to right: $G_0 =  O(isat\text{ } robot1\text{ } ballposition)$, $G_1 = G_0 \land O(goalscored)$, $G_2 = G_{SPL} = G_1 \land O(ballsafe\text{ } S\text{ } isat \text{ } robot1 \text{ } ballposition)$. Rectangular boxes contain fluents for the overlapped branch.}

\end{figure}

\paragraph{\textbf{Reaching the ball.}} First, only the ball is made available by the conceptual layer. The goal is $G_0 = O(isat\text{ }robot1\text{ }ballposition)$, where the PLTL$_f$ operator $O(\cdot)$ ("\emph{once}") was used to require that any generated trace satisfies the predicate "$isat\text{ }robot1\text{ }ballposition$" at least once. The resulting policy, shown in Fig. \ref{fig:policy1}, features only one action that moves the robot to the ball position.

\paragraph{\textbf{Scoring a goal.}} At the conceptual layer, "\emph{Score a goal}" and "\emph{Score in the Opposite Goal}" are different (in the first case two goal posts are required to identify a goal, while in the second case four goal posts are required to identify two different goals and discern which one is the opponent goal), while at the decision-making level, the policy will be the same. The conceptual layer will provide at runtime the correct fluents to determine if a goal is available and the grounded position of the opponent goal. The resulting goal $G_1 = G_0 \land O(goalscored)$ leads to a more complex policy: the additional PLTL$_f$ rule $O(goalscored)$, requires that the goal is scored at least once along any generated trace. Given that the policy enumerates all possible combinations of predicates representing runtime fluents, only a single branch of the policy is show in Fig. \ref{fig:policy2}.

\paragraph{\textbf{Scoring a goal on the SPL field.}} In case the full field is perceived (including the field lines), it is safer for the agent to assume the presence of the opponents on the field, adopting a safer strategy in order to ensure that a goal is scored. The goal is $ G_2 = G_{SPL} = G_1 \land (ballsafe\text{ }S\text{ }isat\text{ }robot1\text{ }ballposition)$, where the additional PLTL$_f$ construct $(ballsafe\text{ }S\text{ }isat\text{ }robot1\text{ }ballposition)$ requires the traces to keep the predicate "$ballsafe$" verified in all states in the states following the state in which the predicate $isat\text{ }robot1\text{ }ballposition$ is verified for the first time: the agent will execute domain actions that keep the ball protected (these actions have the predicate $ballsafe$ in their post-conditions). Result is partially shown in Fig. \ref{fig:policy3}.

The webpage \url{sites.google.com/diag.uniroma1.it/play-everywhere} contains videos and images of NAO robots playing soccer in the wild. The page contents will be extended in the future with novel use-cases.

\section{Conclusions and Future Directions}
\label{sec:conclusion}




In this paper, we have presented a novel approach to robot soccer, which enables the robot to adapt its behavior and goals to the semantics of the environment in real-time. The proposed approach is based on a hierarchical representation of soccer and uses Pure-Past LTL rules to express temporal goals on finite traces. 

Our method aims at being effective in unstructured and dynamic environments, where traditional hard-coded approaches fail. We have demonstrated its effectiveness through three different use-cases, conducted in different scenarios. Experimental results show that the proposed method can adapt to changing conditions and generalize to new and unseen situations, making it highly adaptable to different domains and tasks (see the webpage \url{sites.google.com/diag.uniroma1.it/play-everywhere}).
Overall, our approach represents a significant step towards the development of robots that can perform complex tasks in dynamic and unstructured environments, which is an open research problem in the field of robotics.
As future work, we intend to use an image segmentation system to extract the elements of the field, thus achieving a more complete understanding of the playing environment.

\section*{Acknowledgement}
We acknowledge partial financial support from PNRR MUR project PE0000013-FAIR.

%
%
%
\bibliographystyle{splncs04}
\bibliography{biblio}

\begin{thebibliography}{10}
\providecommand{\url}[1]{\texttt{#1}}
\providecommand{\urlprefix}{URL }
\providecommand{\doi}[1]{https://doi.org/#1}

\bibitem{gamestrategies}
Antonioni, E., Suriani, V., Riccio, F., Nardi, D.: Game strategies for physical
  robot soccer players: A survey. IEEE Transactions on Games  \textbf{13}(4),
  342--357 (2021)

\bibitem{bastianelli-bloisi-gemignani}
Bastianelli, E., Bloisi, D.D., Capobianco, R., Cossu, F., Gemignani, G.,
  Iocchi, L., Nardi, D.: On-line semantic mapping. In: ICAR. pp.~1--6 (2013).
  \doi{10.1109/ICAR.2013.6766501}

\bibitem{Capobianco2015}
Capobianco, R., Serafin, J., Dichtl, J., Grisetti, G., Iocchi, L., Nardi, D.: A
  proposal for semantic map representation and evaluation. In: Mobile Robots
  (ECMR), 2015 European Conference on. pp.~1--6. IEEE (2015)

\bibitem{de2021fondpltlf}
De~Giacomo, G., Favorito, M., Fuggitti, F.: Planning for temporally extended
  goals in pure-past linear temporal logic: A polynomial reduction to standard
  planning  (04 2022)

\bibitem{de2018automata}
De~Giacomo, G., Rubin, S.: Automata-theoretic foundations of fond planning for
  ltlf and ldlf goals. In: IJCAI. pp. 4729--4735 (2018)

\bibitem{de2013LTLsynthesis}
De~Giacomo, G., Vardi, M.Y.: Linear temporal logic and linear dynamic logic on
  finite traces. In: Proceedings of the Twenty-Third International Joint
  Conference on Artificial Intelligence. p. 854–860. IJCAI '13, AAAI Press
  (2013)

\bibitem{kurach2020google}
Kurach, K., Raichuk, A., Sta{\'n}czyk, P., Zaj{\k{a}}c, M., Bachem, O.,
  Espeholt, L., Riquelme, C., Vincent, D., Michalski, M., Bousquet, O., et~al.:
  Google research football: A novel reinforcement learning environment. In:
  Proceedings of the AAAI Conference on Artificial Intelligence. vol.~34, pp.
  4501--4510 (2020)

\bibitem{mynd}
Mattmüller, R., Ortlieb, M., Helmert, M., Bercher, P.: Pattern database
  heuristics for fully observable nondeterministic planning. Proceedings of the
  International Conference on Automated Planning and Scheduling
  \textbf{20}(1),  105--112 (May 2021). \doi{10.1609/icaps.v20i1.13408}

\bibitem{musumeci}
Musumeci, E., Suriani, V., Antonioni, E., Nardi, D., Bloisi, D.: Adaptive team
  behavior planning using human coach commands (07 2022)

\bibitem{pronobis2012large}
Pronobis, A., Jensfelt, P.: Large-scale semantic mapping and reasoning with
  heterogeneous modalities. In: 2012 IEEE international conference on robotics
  and automation. pp. 3515--3522. IEEE (2012)

\bibitem{BHumanCodeRelease2021}
R{\"o}fer, T., Laue, T., Bahr, N., Jaeger, J., Knychalla, J., Lorenzen, T.,
  Matschull, N., Meinken, Y., Monnerjahn, L.M., Plecher, L., Reichenberg, P.:
  {B}-{H}uman team report and code release 2021 (2021), only available online:
  \url{http://www.b-human.de/downloads/publications/2021/CodeRelease2021.pdf}

\end{thebibliography}

\end{document}